\documentclass[letterpaper, 10 pt, conference]{ieeeconf}

\IEEEoverridecommandlockouts
\usepackage{graphics} 
\usepackage{epsfig} 
\usepackage{mathptmx} 
\usepackage{times} 
\usepackage{amsmath} 
\usepackage{amssymb}  
\usepackage{cite}

\newcommand\degree{^\circ}

\title{\LARGE \bf
Design and Stiffness Analysis of a Bio-inspired Soft Actuator with Bi-direction Tunable Stiffness Property
}

\author{Jianfeng Lin, \textit{Student Member, IEEE}, Ruikang Xiao, and Zhao Guo, \textit{Member, IEEE}
\thanks{Research supported by the National Key Research and Development Program of China (No. 2023YFE0202100), the National Natural Science Foundation of China (Grant No. 51605339), and the Key Research and Development Program of Hubei Province (Grant NO. 2020BAB133). We are grateful to the Wuhan University Student Engineering Training and Innovation Practice Center for the venue and equipment support.}
\thanks{J. Lin, R. Xiao, and Z. Guo are at the School of Power and Mechanical Engineering, Wuhan University, Wuhan, 430072, China. Corresponding author to Zhao Guo (e-mail: {\tt\small guozhao@whu.edu.cn})
}%
}%

\begin{document}

\maketitle
\thispagestyle{empty}
\pagestyle{empty}

\begin{abstract}
Modulating the stiffness of soft actuators is crucial for improving the efficiency of interaction with the environment. However, current stiffness modulation mechanisms are hard to achieve high lateral stiffness and a wide range of bending stiffness simultaneously. Here, we draw inspiration from the anatomical structure of the finger and propose a bi-directional tunable stiffness actuator (BTSA). BTSA is a soft-rigid hybrid structure that combines air-tendon hybrid actuation (ATA) and bone-like structures (BLS). We develop a corresponding fabrication method and a stiffness analysis model to support the design of BLS. The results show that the influence of the BLS on bending deformation is negligible, with a distal point distance error of less than 1.5 mm. Moreover, the bi-directional tunable stiffness is proved to be functional. The bending stiffness can be tuned by ATA from 0.23 N/mm to 0.70 N/mm, with a magnification of 3 times. The addition of BLS improves lateral stiffness up to 4.2 times compared with the one without BLS, and the lateral stiffness can be tuned decoupling within 1.2 to 2.1 times (e.g. from 0.35 N/mm to 0.46 N/mm when the bending angle is 45 deg). Finally, a four-BTSA gripper is developed to conduct horizontal lifting and grasping tasks to demonstrate the advantages of BTSA.

\end{abstract}


\section{INTRODUCTION}

   \begin{figure}[thpb]
      \centering
      \includegraphics[scale=0.95]{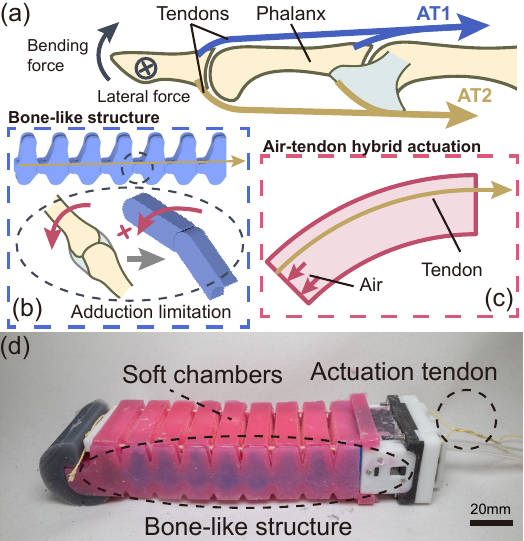}
      \caption{The design concept of BTSA. (a) The anatomical feature of finger. (b) Description of adduction limitation of bone-like structure. (c) Description of air-tendon hybrid actuation. (d) The prototype of BTSA.}
      \label{fig1}
   \end{figure}

   \begin{figure*}[thpb]
      \centering
      \includegraphics[scale=0.9]{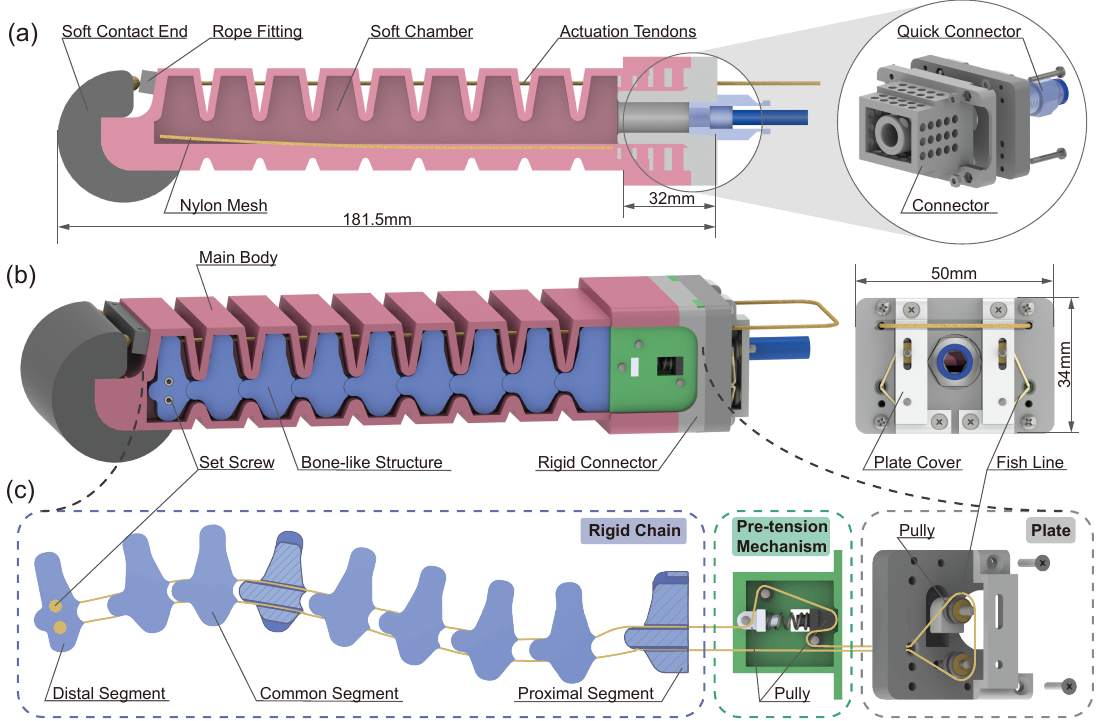}
      \caption{The design details of the BTSA. (a) The sectional view of the main body and the details of the soft-rigid connection. (b) The overall structure (left) and back view (right) of BTSA. (c) The details of the bone-like structure.}
      \label{fig2}
   \end{figure*}
   
Due to the adaptability, soft robotics have gained increasing attention in manipulation\cite{jiang2021hierarchical,sozer2020pressure,chen2021soft}, rehabilitation \cite{polygerinos2015soft,gu2021soft}, grasping\cite{liu2021soft,abondance2020dexterous,chen2017reconfigurable,park2018development,li2022stiffness,wang2022data,gilday2023sensing}, and locomotion\cite{zhang2021inchworm,rafsanjani2018kirigami}. On top of that, tunable stiffness further enhanced the performance by enabling large output force and stability while maintaining compliance and safety\cite{manti2016stiffening}. 

In this framework, active and semi-active stiffness regulation methods were proposed\cite{manti2016stiffening}. The active method is based on antagonistic actuation\cite{robertson2017soft,she2016modeling,manti2015bioinspired,li2017stiffness} like fluid actuation\cite{robertson2017soft}. Varying stiffness by modulation of material property is the semi-active method\cite{majidi2010tunable,shintake2015variable,huynh2022soft,jiang2019chain,jadhav2022variable,wei2016novel} like material jamming\cite{shintake2015variable,huynh2022soft,jiang2019chain,jadhav2022variable,wei2016novel}. However, there is still a gap for the soft actuator to have high lateral stiffness and large bending stiffness modulation range. As shown in Fig. 1a, the bending and lateral stiffness refer to the resistance to deformation by force on bending plane and force perpendicular to the bending plane, respectively. 

For example, Jiang et al.\cite{jiang2019chain} designed a chain-like granular jamming mechanism to enhance stiffness by controlling the pulling force to the mechanism. Although the lateral stiffness can be ensured due to the thick jamming layer, it shows limited bending stiffness modulation range - little distal point position changed under different pulling forces. Contrary to that, the gas-ribbon-hybrid driven actuator proposed by Zhang et al.\cite{zhang2022gas} shows a good compliance, but the lateral stiffness is small because the soft chamber and ribbon cannot provide an enhancement of lateral stiffness. Due to the low lateral stability, the soft gripper cannot lift objects horizontally like rigid grippers, which has a high grasping robustness according to \cite{Makita2019HomogeneousQM}. 

In other studies\cite{sozer2020pressure, hinges2019, zhu2023bioinspired}, mechanical hinges were selected as an internal skeleton to reinforce lateral stiffness. However, the fixed connections limit the overall softness of the actuators. The micropump-activated jamming method\cite{huynh2022soft} that demonstrates high stability can only become stiff at a single bending angle.

Therefore, in this paper, we propose a bi-directional tunable stiffness actuator (BTSA), as shown in Fig.1d. The bi-direction tunable stiffness property enables the actuator with high lateral stiffness and a wide range of bending stiffness. The BTSA consists of bone-like structure (BLS, Fig. 1b) and air-tendon hybrid actuation (ATA, Fig. 1c). The BLS draws inspiration from the morphological feature of the phalanx, simplifying the tuft and the base of phalanx into a semi-column. As Fig. 1b displays, all the segments which are connected by a rope can rotate freely while enhancing the lateral stiffness by the adduction limitation. By changing the pulling force to the rope, the lateral stiffness can be modulated decoupling. The air-tendon hybrid actuation (Fig. 1c) mimics antagonistic muscles of finger (AT1 and AT2 in Fig. 1a), achieving bending and stiffness control through air and tendon antagonism.

We detail the design and fabrication of BTSA. A simplified model for the lateral stiffness is presented. The design is further validated by the influence of the BLSs on deformation, bending and lateral stiffness, and grasping applications.

\section{STRUCTURE DESIGN AND FABRICATION}

\subsection{Structure design}
   
     \begin{figure}[thpb]
      \centering
      \includegraphics[scale=0.95]{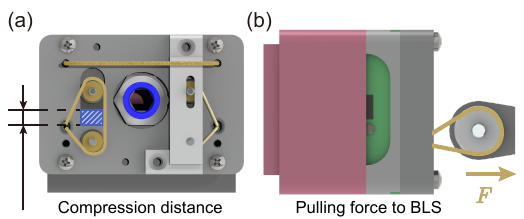}
      \caption{Two methods to control the stiffness of BLS. (a) Add the PLA block (blue) to control the compression distance. (b) Connect the end of the rope to a pully and then control the pulling force to BLS directly.}
      \label{fig3}
   \end{figure}

The main structure of BTSA, as Fig. 2b shows, is composed of three major parts, the main body, bone-like structures (BLSs), and the rigid connector. The two BLSs are attached to each side of the main body, connecting to the rigid connector.

    \begin{figure}[thpb]
      \centering
      \includegraphics[scale=0.95]{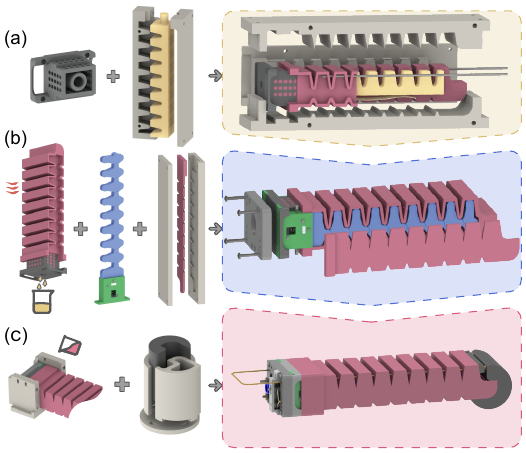}
      \caption{The fabrication process of the proposed BTSA. (a) Fabricating the soft chamber. (b) Fit the BLSs. (c) Assemble other components.}
      \label{fig4}
   \end{figure}
   
Fig. 2a displays the sectional view of the main body and the details of the soft-rigid connection. The main body includes three components: a soft contact end, the soft chamber, and the actuation tendons. The soft chamber (Dragon Skin 20, Smooth-On) is fabricated with a nylon mesh inserted to limit the elongation. The contact pad (Dragon Skin 30, Smooth-On) is designed to increase the fingertip force, glued on the front of the soft chamber\cite{abondance2020dexterous}. The actuation tendons (Kevlar) cross through the holes in both sides of the actuator and tie to the rope fittings at the end of the actuator. These holes’ centers are in the same plane as the symmetrical plane of BLS, which do not go through the chamber. The back of the soft chamber is cast into the connector. The quick connector is connected to the rigid connector through a pipe thread.
   
The design details of the bone-like structures are shown in Fig. 2c. There are three major parts in each BLS, rigid chain, pre-tension mechanism, and plate, connected through a fishline goes across them. The fishline is fixed by two set screws in the distal segment and its path is shown in Fig. 2c in yellow. The bulge and concavity of the each segment are semi-cylindrical, and the holes are perforated along the tangent of the cylinder. The segments are designed as a bone to match the size of the actuator. The pre-tension mechanism (nylon) is to offer the pre-tightening force to BLS. The two pulleys (nylon) on each side of BLS provide a path to guide the fishline and form a closed-loop path. One of the pulleys is connected to the slider (PLA) placed in the chute of the backward plate (PLA). 

As Fig. 3 shows, the lateral stiffness of BLS can be modulated in two ways: the passive way (Fig. 3a) by adding the blocks to change the position of the slider to regulate the pre-tightening force to BLS and the active way (Fig. 3b) by controlling the pulling force to BLS directly. 

\subsection{Fabrication}

 \begin{figure}[thpb]
      \centering
      \includegraphics[scale=0.96]{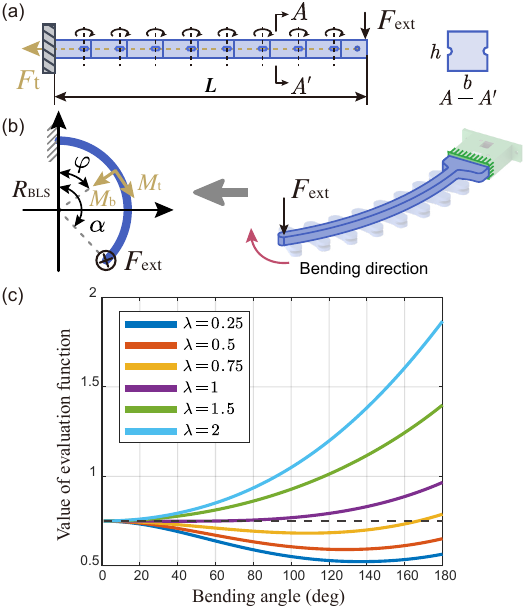}
      \caption{Description of stiffness analysis of BLS. (a) Simplified schematic of BLS. (b) Simplified model of the curved cantilever beam. (c) The trend of stiffness with bending angle for different aspect ratios.}
      \label{fig5}
   \end{figure}

Firstly, fabricate the soft chamber. As Fig. 4a shows, the demolded wax core was fitted to the connector through the cylinder hole. The fluid silicone was poured into the pre-closed mold with the wax core, connector, and an embedded nylon mesh. Then, the soft chamber with a wax core in it was obtained after it was cured. In this case, the soft chamber would be one-piece molding and the porous sidewall of the connector reinforces the airtightness in the soft-rigid-connection boundary.
   
Then, fit the BLSs. As Fig. 4b displays, the soft chamber with a wax core was heated in the water bath to 85℃ for 15 minutes. Squeezed out all the molten wax from the cylinder hole, the soft chamber was assembled with two BLSs and plate. Subsequently, two cover layers were cast and bonded with the soft chamber by uncured silicone.

In the end, assemble other components. As presented in Fig. 4c, the sealing layer of BTSA and the contact end were fabricated. After connecting the quick connector, gluing the contact end, and fitting the actuation tendon, the BTSA was fabricated.

\section{STIFFNESS ANALYSIS}

\subsection{Bending stiffness} 

When the tendon in a relaxed, the actuator driven by pressuring will remain the high compliance. When the tightened tendons limits the bending of BTSA while the air pressure is still rising, The increased gas pressure will improve the bending stiffness.

\subsection{Lateral stiffness}

When BLS works normally, as Fig. 5a displays, the BLS can be considered as a multi-segment freely rotatable cantilever beam. The cross section of the beam is equivalent to the cross section of the bulge of segments. Adjacent segments are tightly bonded under the pulling force $F_{\rm{t}}$ of fishline. 

When the destabilization does not occur, we assumed the pitch between each segment is small enough to be negligible. Hence, the BLS is modeled as a curved cantilever beam curving as the arc of a circle with centre angle $\alpha$ and radius $R_{\rm{BLS}}$ as Fig. 5b shows. The centre angle here is equal to the bending angle of BTSA. The axial length of the BLS is constant, so $R_{\rm{BLS}}\alpha =L$. 

For this equivalent beam, the contribution of shear force to the strain energy can be neglected. Therefore, the strain energy $U_{BLS}$ satisfying the following equation:
$$
U_{BLS}=\frac{1}{EI}\int_0^{\alpha}{M_{\rm{b}}^{2}R_{\rm{BLS}}\mathrm{d\varphi}+}\frac{1}{\mathrm{G}I_p}\int_0^{\alpha}{M_{\rm{t}}^{2}R_{\rm{BLS}}\mathrm{d\varphi}}\eqno{(1)}
$$where $M_{\rm{t}}$ and $M_{\rm{b}}$ are torsion moment and bending moment, respectively, $E$ is Young’s modulus, $I$ refers to the area moment of inertia, $G$ is modulus of rigidity, and $I_p$ represents polar moment of inertia.The $G$ and $I_p$ can be obtained from Poisson’s ratio $\nu$ and aspect ratio $\lambda$ of the beam:
$$
\begin{cases}
        G=E/\left[2\left( 1+\nu \right)\right]\\
	I_p=I\left( 1+\lambda ^2 \right)\\
\end{cases}.\eqno{(2)}
$$

According to Castigliano’s second theorem, the stiffness can be derived as
$$
k=\frac{4EI}{C^3}F\left( \alpha \right)\eqno{(3)} 
$$
$$
F\left( \alpha \right) ={{1}\Bigg/{\left[ A_{bending}\left( \alpha \right) +\frac{2\left( 1+\nu \right) A_{torsion}\left( \alpha \right)}{\left( 1+\lambda ^2 \right)} \right]}}.\eqno{(4)} 
$$$A_{bending}\left( \alpha \right)$ and $A_{torsion}\left( \alpha \right)$ only vary with centre angle $\alpha$, representing the influence of bending and torsion on displacement. Hence, $F\left( \alpha \right)$ is considered as an evaluation function to analyze the stiffness variation with centre angle $\alpha$ and aspect ratio $\lambda$.

The values of evaluation function are shown as Fig. 5c with that aspect ratio varies from 0.25 to 2 and Poisson's ratio is about 0.35\cite{PLA2005PR}. From Fig. 5c, the lateral stiffness increases with the bending when the aspect ratio is greater than 1. This is a desired property to ensure the lateral stability during bending. Meanwhile, according to euqation (3), larger geometrical sizes means higher lateral stiffness. However, these parameters are limited by the size of soft chamber. Excessive parameters can cause undisered axial deformation. As a result, we chose 1 as the aspect ratio of BLS.

\begin{figure}[thpb]
      \centering
      \includegraphics[scale=0.96]{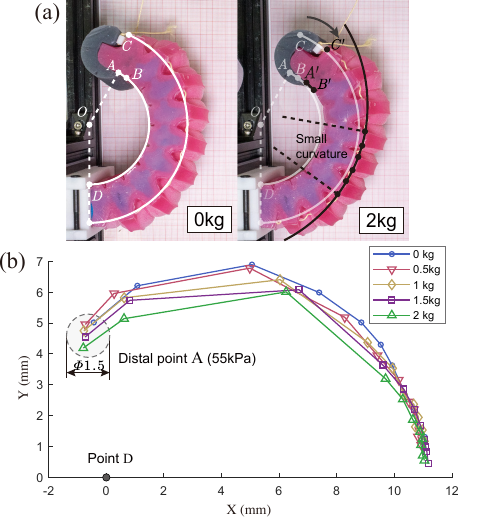}
      \caption{Characterization of the influence of BLS on bending. (a) An intuitive comparison of bending between when the pulling force to BLSs is 0kg and 2kg. (b) Results of position of point A under pulling forces.}
      \label{fig6}
   \end{figure}


\section{EXPERIMENTS AND RESULTS}

\subsection{The influence of BLS on bending}

     \begin{figure*}[thpb]
      \centering
      \includegraphics[scale=0.98]{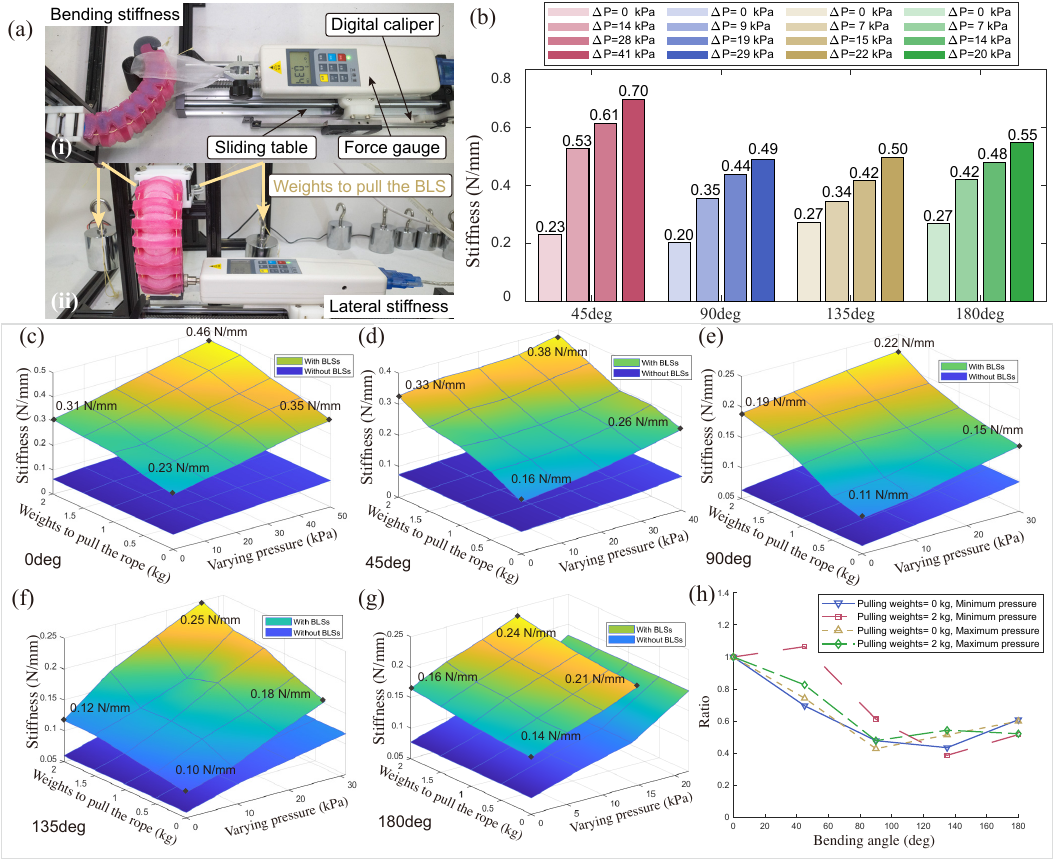}
      \caption{The experimental platforms and results of stiffness testing. (a) The experiment platforms for (i) bending stiffness and (ii) lateral stiffness. (b) The testing results of bending stiffness. The testing results of lateral stiffness when bending angle is (c) 0 deg, (d) 45 deg, (e) 90 deg, (f) 135 deg, and (g) 180g. (h) The change of stiffness ratio with bending angle.}
      \label{fig7}
   \end{figure*}

Due to the friction between each segment, the BLS would might not work ideally that provides only the lateral stability enhancement but shows little influence on deformation. To test such influence, the second method (Fig. 3b) was chosen to control the lateral stiffness. Fig. 6a shows an intuitive comparison. When the pulling weight increased to 2 kg, the deformation was composed of several different segmented curvatures with a smaller curvature in the middle. This is different with the normal soft-actuator deformation with constant curvature like when pulling weight was 0 kg\cite{zhang2022gas,liu2021soft,low2020bidirectional}.

Further, we tested the position of distal point A to quantify this influence. BTSA was pressurized from 0 to 55 kPa with five groups of pulling forces. The center of the minimum circle which was calculated to cover the five recorder points was the average point. As Fig. 6b shows, with the increase of the pulling weights, all groups share the similar trends and the maximum error is within 1.5mm. This error is small enough to be ignored in most application cases.

\subsection{Bending stiffness}

As shown in Fig. 7 a(i), a carbon fiber tube with 26mm diameter was connected to the force gauge. The sliding table was programmed to move 1mm rightward each time until the total distance reached 10mm and the pulling forces during this process would be recorded. For four groups of bending angle, four sets of incremental air pressures were set from the actuation pressure (the pressure when the actuator reaches certain bending angle) to maximum allowable pressure (85 kPa). The bending stiffness was calculated by the incremental ratio of the average force and displacement of three repeated experiments.

Fig. 7b lists all the test results. For all the four different bending angles, the stiffness was enhanced from 1.9 to 3.0 times, with a maximum stiffness of 0.70 N/mm, indicating that the stiffness in the bending direction is tuned successfully through proposed mechanism.

\subsection{Lateral stiffness}

    \begin{figure}[thpb]
      \centering
      \includegraphics[scale=0.98]{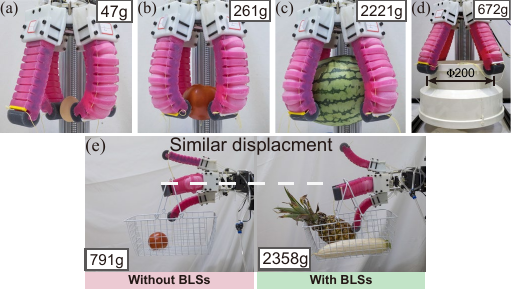}
      \caption{The application cases of BTSA. Successful grasping tasks include (a) an egg, (b) a tomato, (c) a watermelon, and (d) a pipe. (e) Comparison between the gripper with BLSs and without BLSs.}
      \label{fig8}
   \end{figure}

The lateral stiffness was tested as Fig. 7a(ii) illustrates. The force gauge contacted by the end of the side of BTSA. Two weights were connected to each BLS through pulleys to change the pulling force by the second stiffenss modulation method (Fig. 3b). 
   
 For BTSA, the lateral stiffness is influenced by both the input pressure and BLSs. Hence, the experiment has two variations: pulling weights to BLS and varying pressures. Five groups of weights 0kg, 0.5kg, 1kg, 1.5kg and 2kg and at least three groups of pressure at incremental steps of 10 kPa were conducted for five groups of bending angles. For comparison, the lateral stiffness for actuator without BLSs was also tested. The sliding table moved at a 1-mm interval from 0 to 10mm to control the deflection. The stiffness was calculated after three repeated experiments.

All the results are shown from Fig. 7c to 7g. Comparisons between the lateral stiffness of BTSA with BLSs and without BLSs verifies the enhanced stiffness property of BLS. The maximum magnification reaches about 4.2 times when the bending angle is $0\degree$. Meanwhile, for different input pressure, the lateral stiffness can be tuned by BLS within 1.2 to 2.1 times. These results proved that the lateral stiffness can be tuned decoupling successfully. 

In addition, we choose four points in each angle and investigate the variation in lateral stiffness with angle to verify the analytical model. The ratios of stiffness relative to the value when the bending angle is 0 are calculated as Fig. 7h depicts. The graph shows that there has been a slight decrease followed by a rising, which does not match the design goal but is close to the trend of the analysis model when the aspect ratio $\lambda$ is smaller than 1. The simplification of the equivalent beam might contribute to the error since the segments are connected by elastic fishline. 

\subsection{Application cases}

To demonstrate the performance of BLS, a four-BTSA gripper was designed. Each BLS was tightened to about 10N by the first stiffness regulation method (Fig. 3a). Firstly, several cross-weight and cross-scale tasks were successfully conducted, including an egg (47g, Fig. 8a), a tomato (261g, Fig. 8b), a watermelon (2221g, Fig. 8c), and a pipe (672, 200 mm diameters, Fig. 8d). Then, horizontal lifting was performed. The one with tightened BLSs could lift heavier targets (2358g) than the one without BLSs and shared similar displacement. These successful graspings are the results of the bending and lateral stiffness adjustment.

\section{CONCLUSIONS}

In this work, we proposed a novel soft actuator design that can regulate bending and lateral stiffness, inspired by the anatomical structure of the finger. Our design offered a high lateral stiffness and a large range of bending stiffness, and we presented a theoretical stiffness analysis to support the design and analyse lateral stiffness. We also demonstrated that the influence of the rigid structure on deformation is negligible and verified the bi-directional stiffness regulation property through stiffness tests and application cases.




\bibliographystyle{IEEEtran}
\bibliography{IEEEabrv,Reference}

\end{document}